\documentclass{article}

     \PassOptionsToPackage{numbers, compress}{natbib}
\usepackage[final]{neurips}

\usepackage[utf8]{inputenc} % allow utf-8 input
\usepackage[T1]{fontenc}    % use 8-bit T1 fonts
\usepackage{hyperref}       % hyperlinks
\usepackage{url}            % simple URL typesetting
\usepackage{booktabs}       % professional-quality tables
\usepackage{amsfonts}       % blackboard math symbols
\usepackage{nicefrac}       % compact symbols for 1/2, etc.
\usepackage{microtype}      % microtypography
\usepackage{graphicx}
\usepackage{subfiles}
\usepackage{placeins}
\usepackage{caption}
\usepackage{subcaption}
\usepackage{amsmath}
\usepackage{amssymb}
\usepackage{amsthm}
\usepackage{bm}
\usepackage{pifont}
\usepackage{lipsum}  
\usepackage{enumitem}
\usepackage{multirow,tabularx}
\usepackage{makecell}
\usepackage{floatrow}
\usepackage{sidecap}
\usepackage{algorithm}
\usepackage[dvipsnames]{xcolor}
\usepackage{appendix}

\newlist{todolist}{itemize}{2}
\setlist[todolist]{label=$\square$}
\usepackage{pifont}
\hypersetup{
    colorlinks,
    linkcolor={red!70!black},
    citecolor={blue!80!black},
    urlcolor={cyan!50!black}
}

\title{Monitoring snow avalanches from SAR data\\with deep learning}

\author{%
  Filippo Maria Bianchi\\
  UiT the Arctic University of Norway\\
  NORCE, Norwegian Research Centre AS\\
  \texttt{filippo.m.bianchi@uit.no} \\
  \And
  Jakob Grahn \\
  NORCE, Norwegian Research Centre AS\\
  \texttt{jgra@norceresearch.no} \\
}
\usepackage[
    nomain,
    acronym,
    automake,
    stylemods=all,
    style=super
]{glossaries-extra}

\makeglossaries
\setabbreviationstyle[acronym]{long-short}
\glsdisablehyper

\newacronym{s1}{S1}{Sentinel-1}
\newacronym{sar}{SAR}{synthetic aperture radar}
\newacronym{rs}{RS}{remote sensing}
\newacronym{eo}{EO}{earth observation}
\newacronym{cnn}{CNN}{convolutional neural network}

\begin{document}
\maketitle

\begin{abstract}
Snow avalanches present significant risks to human life and infrastructure, particularly in mountainous regions, making effective monitoring crucial. Traditional monitoring methods, such as field observations, are limited by accessibility, weather conditions, and cost. Satellite-borne Synthetic Aperture Radar (SAR) data has become an important tool for large-scale avalanche detection, as it can capture data in all weather conditions and across remote areas. However, traditional processing methods struggle with the complexity and variability of avalanches.
This chapter reviews the application of deep learning for detecting and segmenting snow avalanches from SAR data. Early efforts focused on the binary classification of SAR images, while recent advances have enabled pixel-level segmentation, providing greater accuracy and spatial resolution. A case study using Sentinel-1 SAR data demonstrates the effectiveness of deep learning models for avalanche segmentation, achieving superior results over traditional methods.
We also present an extension of this work, testing recent state-of-the-art segmentation architectures on an expanded dataset of over 4,500 annotated SAR images. The best-performing model among those tested was applied for large-scale avalanche detection across the whole of Norway, revealing important spatial and temporal patterns over several winter seasons.
\end{abstract}

\section{Introduction}
\label{sec:introduction}

Each year, snow avalanches claim more than 250 lives worldwide and cause significant damage to infrastructures, like roads, buildings, and ski areas \cite{fuchs2005, schweizer2021snow}. To mitigate this hazard, strategies such as constructing deflection barriers or triggering controlled avalanches with explosives are commonly used. Although effective, these approaches are costly and can have negative impacts on the environment \cite{jenkins2000evaluating}. Because of these drawbacks, reliable avalanche forecasting is considered one of the most practical and cost-effective ways to reduce both human and economic losses \cite{fuchs2007}. 

Predicting avalanche danger involves assessing snowpack stability. Traditionally, this has been done by digging snow pits and examining the snow structure directly in the field. While this gives valuable insights, it is labor-intensive, time-consuming, and potentially dangerous — especially when avalanche danger is high. Consequently, avalanche forecasters often rely on observations of recent avalanche activity and weather patterns to understand snowpack conditions. 

Historically, observations of avalanche activity were limited to manual field reports, leaving many regions unmonitored. However, advancements in satellite remote sensing have made large-scale monitoring of avalanche activity feasible. In particular, space-borne \gls{sar} offer detailed insights into avalanche debris over vast and remote regions, being agnostic to weather conditions or daylight. Despite some limitations, SAR provided a broader and more consistent tracking of avalanche activity, greatly improving our understanding of snowpack dynamics across entire mountain ranges \cite{eckerstorfer2015manual}.

\subsection{SAR - an (almost) ideal sensor for avalanche mapping}

\begin{figure}
    \centering
    \includegraphics[width=1.0\linewidth]{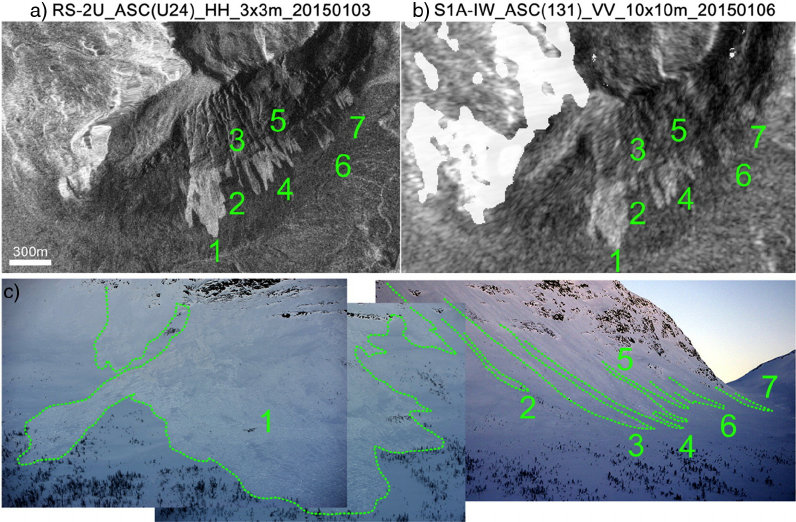}
    \caption{Observations of avalanche debris using satellite-based SAR data. (a) Radarsat-2 ultrafine mode backscatter image. (b) Sentinel-1A IW mode backscatter image, acquired three days later. (c) Oblique photographs showing the avalanche debris are outlined in green for reference. Source: \cite{eckerstorfer2016remote}.}
    \label{fig:fig0}
\end{figure}

\gls{sar} is an active remote sensing technology that transmits coherent microwave pulses toward the Earth's surface and measures the backscattered signal. Since these pulses are coherent, their relative phase information can be used to achieve \textbf{high spatial resolution} — typically down to just a few meters. This enables \gls{sar} to detect detailed features such as avalanche debris. Additionally, because \gls{sar} generates its own signal, it can operate \textbf{independently of daylight}, making it particularly valuable for monitoring avalanches, which often occur during poor visibility or limited daylight conditions. Most \glspl{sar} operate at frequencies that are \textbf{unaffected by atmospheric conditions}, ensuring consistent and reliable data even during challenging weather.

Figure \ref{fig:fig0} shows several avalanches captured using two different satellites: Radarsat-2 and Sentinel-1. The bright areas represent high backscatter values, where avalanche debris clearly stands out compared to the surrounding snow. The increased backscatter is primarily attributed to the higher surface roughness and the compaction of snow within the debris.

Despite having properties that are ideal for avalanche monitoring, there are some limitations to using \gls{sar} for this purpose. Firstly, unless the frequency is relatively high (e.g.\, X-band or higher), microwaves are largely insensitive to dry, low-density snow. Under these conditions, avalanche debris may appear almost transparent, making detection challenging \citep{eckerstorfer2022performance}. This limitation means that \gls{sar}-based detections might be \textbf{biased towards compact or wet snow} avalanches, while some loose and dry snow avalanches could go unnoticed.  The specific conditions under which this occurs are not yet well understood. 

Secondly, although \gls{sar} provides high spatial resolution, it can suffer from a relatively \textbf{low temporal resolution}, due to the time interval between image acquisitions. The interval depends on the satellite’s orbit and the number of satellites in the constellation. For instance, the Sentinel-1 constellation has a 6-day repeat-pass cycle. If an avalanche occurs early in this interval, changing weather conditions—such as precipitation, wind, or temperature fluctuations—may alter the snowpack enough to make the debris difficult to detect by the next pass. However, in areas where multiple orbits overlap, this time interval is significantly reduced. This overlap is more pronounced at higher latitudes, like in Northern Norway, where up to 8 overlapping orbits cover the same region, reducing the effective time interval between acquisitions to less than a day.

Thirdly, \gls{sar} cannot effectively image very steep slopes that either face directly toward or away from the radar due to \textbf{layover} and \textbf{shadowing} effects. Since steep slopes are typical release zones for avalanches, these effects can limit the ability to capture the entire avalanche path. However, as avalanche debris typically accumulates where the slope flattens out, \gls{sar} can still provide valuable information for post-event analysis and avalanche mapping.

So, while \gls{sar} has certain limitations that must be considered when using it for avalanche mapping and forecasting, it offers a view of snow avalanche dynamics at spatial and temporal scales that were previously impossible. 

\subsection{The advent of Deep Learning in Remote Sensing for disaster management}

The advent of deep learning, particularly Convolutional Neural Networks (CNNs), has transformed the field of remote sensing by offering innovative approaches to analyzing complex and high-dimensional data. CNNs and other deep learning models have proven highly effective across various remote sensing tasks, such as land cover classification~\cite{alem2020deep}, change detection~\cite{luppino2021deep}, and vegetation monitoring~\cite{gazzea2021automated}. These models excel at automatically extracting and learning relevant features from vast datasets, thanks to their capacity to handle large volumes of data and capture hierarchical spatial features—an essential capability for interpreting intricate patterns found in satellite imagery~\cite{geng2020multi}.

A significant advantage of deep learning in remote sensing is its ability to generalize across different datasets and sensor types. CNNs have been effectively adapted to handle diverse inputs, from high-resolution optical images to radar data, demonstrating remarkable flexibility in addressing various challenges across multiple domains~\cite{luppino2021deep}. This ability to simultaneously process both spectral and spatial information enables a more nuanced analysis of remote sensing data, far surpassing the capabilities of conventional techniques. The adaptability of deep learning has paved the way for integrating data from multiple sources, resulting in more robust and comprehensive disaster monitoring systems~\cite{linardos2022machine}.

In the context of disaster management and environmental monitoring, deep learning models have demonstrated significant advancements over traditional methods, offering improved accuracy, speed, and automation in critical tasks such as flood detection~\cite{wu2023near}, wildfire monitoring~\cite{tong2024real}, and oil spill detection~\cite{bianchi2020large}. 
These capabilities are particularly crucial for a near-real-time analysis of extensive satellite data, enabling effective monitoring and timely responses to hazardous events such as snow avalanches.

\section{History of regional avalanche monitoring with SAR}

Satellite-borne SAR has introduced a new way to monitor avalanches on a large scale, providing coverage that ground-based methods could never achieve. However, this technology generates a vast amount of data. In Norway, for example, avalanche-prone terrain extends over 1,750 km from south to north, and Sentinel-1 captures imagery at a 10-meter resolution, resulting in enormous datasets that are impractical to analyze manually.

Moreover, human interpretation can introduce inconsistencies and biases, as results often vary depending on the analyst’s experience and judgment. Automatic methods offer a more consistent and objective approach, reducing biases and ensuring reliable results across large regions. Early methods faced challenges in processing these datasets effectively, but advancements in machine learning and deep learning greatly advanced the accuracy and efficiency in the analysis of \gls{sar} data.

\subsection{Early methods}

Early efforts to detect avalanches in SAR data relied on standard signal processing techniques, such as segmentation based on thresholding. Especially in dynamic and complex terrains, these methods frequently failed to distinguish avalanche debris from other snow-covered surfaces with similar radar signatures due to the highly variable nature of snow, leading to numerous false alarms. These methods were also sensitive to noise and the choice of thresholds, which rely on simplistic statistical assumptions that cannot fully capture the complexity of avalanche events. Consequently, substantial features engineering and manual corrections are needed, making the process labor-intensive, inefficient, and prone to errors~\cite{malnes2015first}.

Another key limitation of these early methods was their focus on pixel-level information without considering broader contextual features, such as the shape, texture, and spatial patterns of avalanche debris. Furthermore, local topographical factors like slope angle and aspect—crucial for understanding avalanche dynamics—were mostly disregarded, except for masking areas where avalanches were unlikely to occur~\cite{vickers2016method}.

\subsection{Binary Classification of Avalanches with Deep Learning}

One of the earliest applications of deep learning for snow avalanche monitoring focused on binary classification of SAR images, where CNNs were used to classify fixed-size patches of Sentinel-1 radar data into two categories: those containing at least one avalanche and those without any avalanche~\cite{kummervold2018avalanche, hamar2016automatic}.

\begin{figure}
    \centering
    \includegraphics[width=0.5\linewidth]{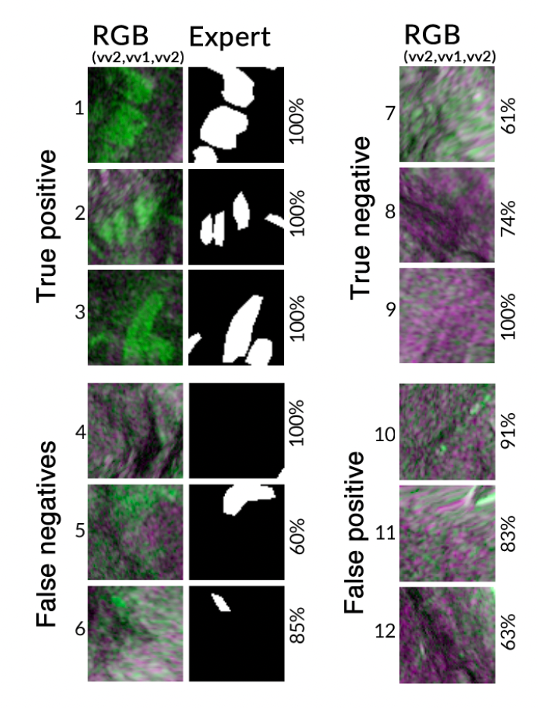}
    \caption{Examples of slices classified by the CNN and an expert (1-6) as well as false detections (7-12) The percentages indicate the network’s confidence that the image is an avalanche (1-6) or that the image is not an avalanche (7-12). Source: \cite{kummervold2018avalanche}.}
    \label{fig:fig1}
\end{figure}

In particular, the study in \cite{kummervold2018avalanche} trained CNNs to detect avalanche features within patches of SAR images. While the model achieved reasonable accuracy, the performance was highly dependent on the size of the image patches used. Larger patches covered more area, making it easier to detect the presence of at least one avalanche but often at the expense of spatial precision, as this approach tended to overestimate the affected areas. In contrast, smaller patches provided finer spatial resolution but were more prone to missing avalanches, leading to higher false negative rates (see Figure~\ref{fig:fig1}).

Although this patch-wise classification represented a step forward in leveraging deep learning for avalanche detection, it had several notable limitations. The patch size heavily affected the model behavior: larger patches increased the likelihood of detecting avalanches but compromised spatial accuracy, while smaller patches improved resolution but introduced misclassification issues. Additionally, the model was only capable of providing a binary outcome for each patch, which limited its ability to capture the presence of multiple avalanches within a single area. This lack of detailed information reduced the effectiveness of the model, particularly in complex terrains where multiple avalanches could occur simultaneously.

Despite these challenges, the initial efforts in binary classification laid important groundwork for further advancements. These early models highlighted the need to balance detection accuracy with spatial resolution and inspired subsequent research aimed at developing more detailed methods for avalanche detection, such as those that could identify the precise location, size, and shape of avalanches.

\subsection{Recent Advances: segmentation of SAR images with deep learning}

Recent advancements in deep learning have led to significant improvements in the detection and mapping of avalanches from remote sensing data, particularly in the context of segmentation. Unlike earlier methods performing binary classification of image patches, modern approaches focus on pixel-wise segmentation, which aims to classify each pixel in an image, thereby providing a more precise delineation of avalanche boundaries and improving the spatial resolution of detection.

Deep learning architectures such as U-Net~\cite{ronneberger2015u} and its variants have emerged as powerful tools for segmenting snow avalanches in remote sensing images, including SAR. These models are designed to perform dense prediction, enabling them to label each pixel in an image based on learned features that represent avalanche characteristics such as texture, shape, and context~\cite{long2015fully}. Recent studies have demonstrated that these models can effectively capture the complex patterns associated with avalanche debris, even in challenging mountainous terrains~\cite{eckerstorfer2022performance}.

Advanced deep learning models can incorporate contextual information, such as topographical data and multi-temporal images, to improve the accuracy and reliability of segmentation outcomes. Techniques like multi-scale feature extraction, attention mechanisms, and data augmentation have further enhanced the robustness of these models against diverse snow and terrain conditions~\cite{liu2024ava}. One notable development in this area is the application of hybrid models that combine SAR data with other remote sensing modalities, such as optical images or LiDAR, to improve segmentation performance~\cite{kapper2023automated}. 
These approaches leverage the complementary information within different data types to enhance the overall accuracy of avalanche detection and mapping.

Despite these advances, several challenges remain in deploying deep learning models for SAR image segmentation in operational avalanche monitoring. These include the scarcity of labeled data, the need for models to generalize across different geographical regions and SAR sensor types, and the computational complexity associated with training and deploying deep learning models on large datasets.

The seminal work presented in the paper "Snow avalanche segmentation in SAR images with fully convolutional neural networks" by \citep{bianchi2020snow} represents a pivotal contribution to this evolving field. The study will be discussed in detail in the next section, highlighting the methodologies employed, the results obtained, and the potential implications in avalanche monitoring. Then, to conclude the chapter, recent extensions of this work and future directions will be discussed.

\section{Segmentation of snow avalanches with fully convolutional neural networks}

We will now discuss the case study presented in \citep{bianchi2020snow}, which explores the application of deep learning for pixel-level segmentation of snow avalanches using Sentinel-1 SAR imagery. This has been one of the first works introducing a Fully Convolutional Network (FCN) for detecting avalanches at a finer granularity compared to earlier methods that relied on patch-based classification. 
The proposed FCN incorporates novel input features, including the potential angle of reach (PAR), to improve detection accuracy by integrating contextual and topographical information. The model was trained on a large dataset of manually labeled SAR images and achieved an F1 score of 66\%, significantly outperforming the existing state-of-the-art algorithm, which had an F1 score of 38\%. The study demonstrated that the proposed deep learning approach is effective in detecting most avalanches, including some that were overlooked during manual annotation, while only missing a few smaller events.

In the next sections, we present the study in detail, covering the dataset construction, the deep learning architecture, the experimental setup, and the obtained results.

\subsection{Dataset construction}

The dataset used in the study consists of $118$ SAR images acquired from the Sentinel-1 satellites over two mountainous regions in Northern Norway during the period from October 2014 to April 2017. The SAR data were captured in the interferometric wideswath (IW) mode and processed to create ground range detected products. Each image was radiometrically calibrated to radar backscatter (sigma nought) values, downsampled from a resolution of 10 meters to 20 meters, and geocoded onto a $20$-meter resolution UTM grid using a digital elevation model (DEM) with 10-meter resolution. The resulting images were transformed to decibel (dB) values and clipped to a range between $-25$ and $-5$ dB to remove noise and clamp the backscatter values to intervals where avalanches are visible.

\begin{figure}
    \centering
    \includegraphics[width=\linewidth]{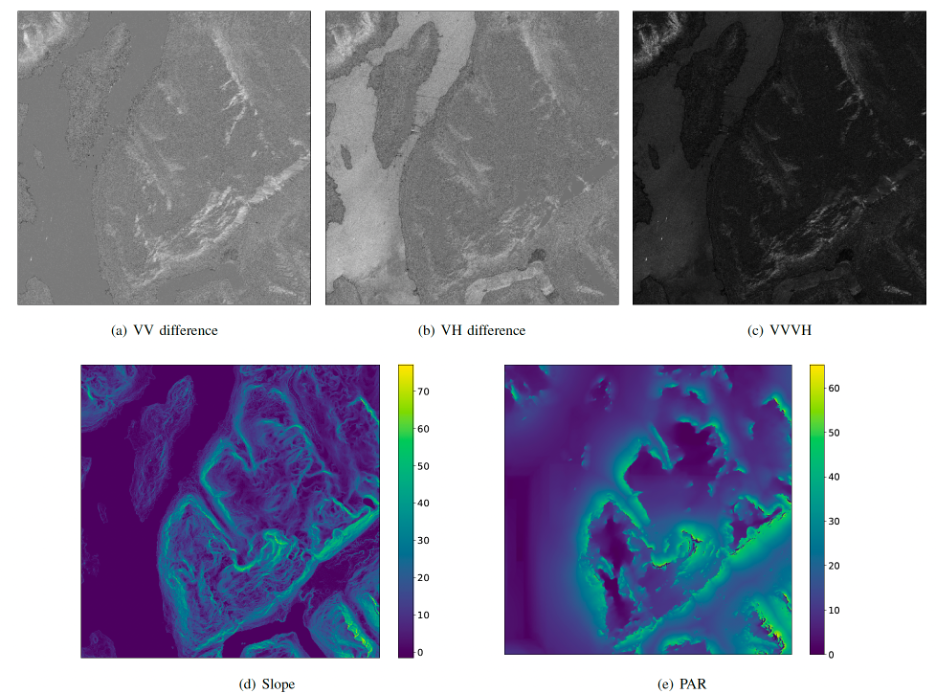}
    \caption{(a, b) the SAR features obtained from the difference in the VV and VH channels. (c) product VVVH of the squared differences. (d, e) slope and PAR feature maps. Only a small area ($1k \times 1k$ pixels) of the actual scene is depicted here. Source: \cite{bianchi2020snow}.}
    \label{fig:fig2}
\end{figure}

Three primary SAR features were generated for use in the deep learning model: (1) the difference in horizontal polarization (VV) between a reference and activity image, (2) the difference in vertical polarization (VH) between these images, and (3) the point-wise product of the squared differences of VV and VH (VVVH). These features were rescaled to the range $[0, 1]$ to enhance the model's ability to detect avalanches (Fig.~\ref{fig:fig2}a-c).

For the labeling process, a human expert manually annotated each image to create a binary segmentation mask, distinguishing avalanche from non-avalanche pixels. 
To identify changes indicative of avalanche activity, the expert utilized a difference of two images acquired at different time steps. The images were created from three channels (R [VV reference], G [VV activity], and B [VV reference]). 
In total, the dataset contained $6,345$ avalanches with approximately $712,945$ pixels labeled as "avalanche" and over $3.6$ billion pixels labeled as "non-avalanche".

\begin{figure}
    \centering
    \includegraphics[width=0.7\linewidth]{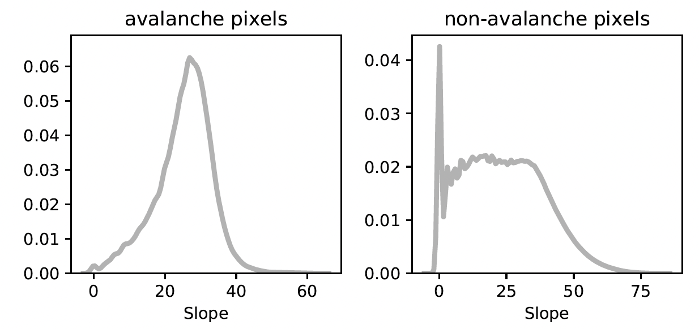}
    \caption{Distribution of the slope angle for avalanche and non-avalanche pixels. Source: \cite{bianchi2020snow}.}
    \label{fig:fig3}
\end{figure}

To further enhance the dataset, two topographical features derived from the DEM were included: the slope angle and the PAR (Fig.~\ref{fig:fig2}d,e). 
The slope angle, computed from the gradient of the DEM, is a critical factor in avalanche detection as avalanches typically initiate on slopes between $35-45$ degrees and deposit on gentler slopes. 
Figure~\ref{fig:fig3} illustrates the distribution of the slope angle for avalanche and non-avalanche pixels, showing that avalanche pixels are mainly concentrated around slopes of $20-35$ degrees.

\begin{figure}
    \centering
    \includegraphics[width=0.3\linewidth]{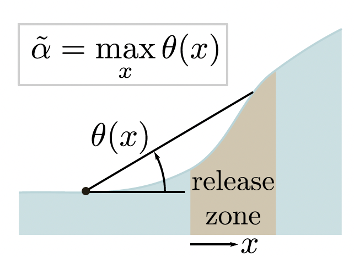}
    \caption{Computation of the potential angle of reach (PAR) $\tilde \alpha$. $\theta(x)$ denotes the angle between the horizontal and the line drawn from a point in a release zone, denoted $x$, to the point of interest. Source: \cite{bianchi2020snow}.}
    \label{fig:fig4}
\end{figure}

\begin{figure}
    \centering
    \includegraphics[width=0.6\linewidth]{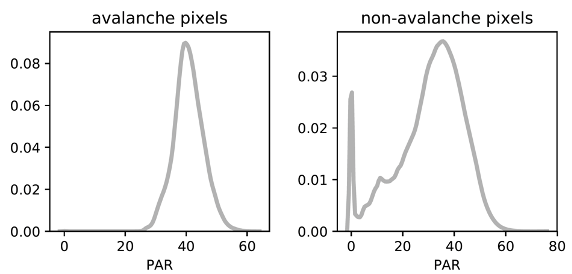}
    \caption{Distribution of the PAR for avalanche and non-avalanche pixels. Source: \cite{bianchi2020snow}.}
    \label{fig:fig5}
\end{figure}

The PAR is a novel introduced feature that estimates the likelihood of an avalanche reaching a specific location, defined as the elevation angle between the furthest avalanche runout point and the highest release point. 
Figure~\ref{fig:fig4} shows the computation of the PAR $\tilde \alpha$.
Unlike traditional runout masks, which filter areas where avalanches are unlikely, the PAR is used as an additional input feature to guide the deep learning model's attention toward areas more likely to contain avalanches. Figure~\ref{fig:fig5} illustrates the distribution of the PAR for avalanche and non-avalanche pixels, hinting that this feature could help to separate the two classes.

\subsection{Deep learning model: architecture and training}

The deep learning model is based on a Fully Convolutional Network (FCN) architecture inspired by the U-Net model. The architecture consists of an encoder-decoder structure designed to perform pixel-level segmentation in Sentinel-1 SAR imagery. The encoder extracts hierarchical features from the input images, progressively reducing the spatial dimensions while increasing the number of filters to capture patterns of increasing complexity. The decoder then reconstructs these features into an output segmentation mask, indicating the avalanche pixels.

\begin{figure}
    \centering
    \includegraphics[width=0.7\linewidth]{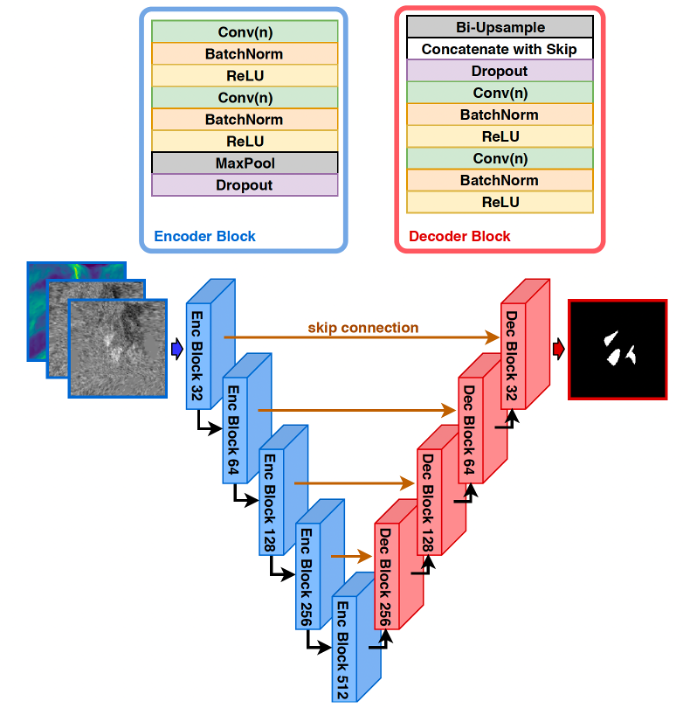}
    \caption{The FCN architecture used for segmentation. Conv(n) stands for a convolutional layer with n neurons. For example, $n = 32$ in the first Encoder Block, $64$ in the second, and so on. Source: \cite{bianchi2020snow}.}
    \label{fig:fig6}
\end{figure}

Fig.~\ref{fig:fig6} illustrates the architecture of the FCN, highlighting both the encoder and decoder blocks. The encoder comprises multiple convolutional layers, each followed by batch normalization, ReLU activation functions, max-pooling layers, and dropout layers. The decoder mirrors this structure, using bilinear upsampling to gradually restore the spatial dimensions. Additionally, skip connections link corresponding layers in the encoder and decoder, allowing the model to recover spatial details lost during downsampling and to improve the pixel-wise alignment of the segmentation output.

\begin{figure}
    \centering
    \includegraphics[width=\linewidth]{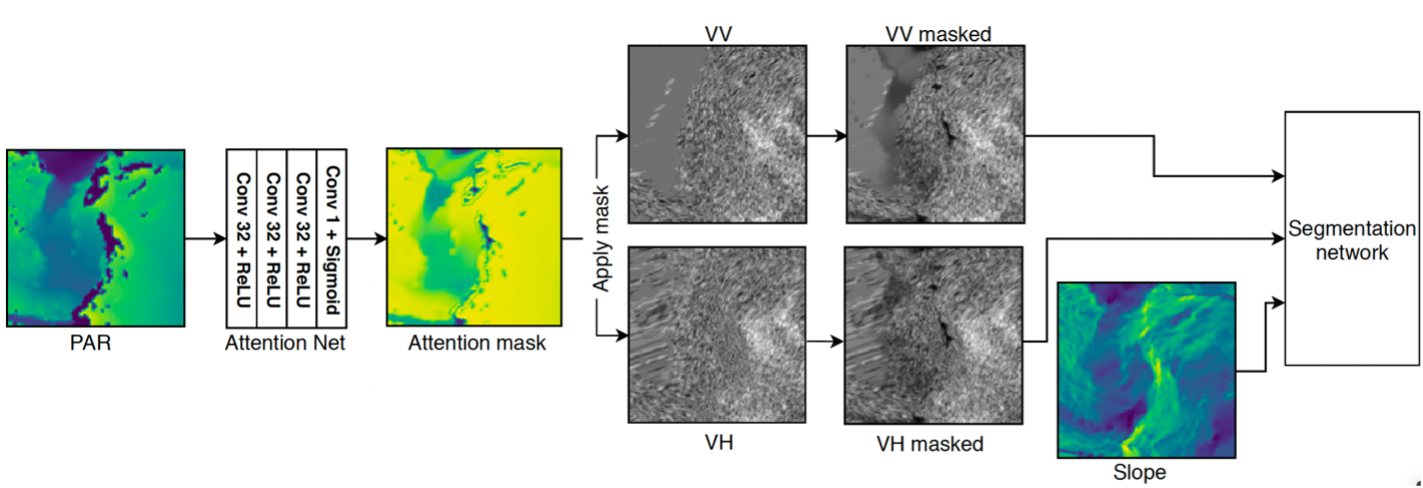}
    \caption{For each patch, the Attention Net generates an attention mask from the PAR features and applies it to the VV, VH, and VVVH SAR features. The masked SAR features and the slope (not masked) are then fed into the FCN. The Attention Net is jointly trained with the FCN by minimizing the segmentation error. Source: \cite{bianchi2020snow}.}
    \label{fig:fig7}
\end{figure}

The proposed model incorporates some novel features to enhance its performance, such as a sub-module (Attention Net) that computes an attention mask conditioned on the PAR.
The mask is applied element-wise to the input SAR features (VV, VH, and VVVH channels) before they are fed into the segmentation network (see Fig.~\ref{fig:fig7}). This helps the model focus on critical regions within the input images, improving segmentation accuracy.

To address the issue of class imbalance — where avalanche pixels are significantly underrepresented in the dataset — the loss function is designed to give more importance to misclassified avalanche pixels. 
After experimenting with various loss functions, a binary cross-entropy loss with class balancing outperformed other loss functions such as the Jaccard-distance loss and the Lovász-Softmax loss.
The network was trained on small patches ($160 \times 160$ pixels) instead of the entire image to mitigate memory constraints and to introduce stochasticity that regularizes the learning process. The training process utilized data augmentation techniques, such as random horizontal and vertical flips, rotations, shifts, zooming, and shearing, to prevent overfitting and enhance the model's ability to generalize to new data. To refine the predictions at inference time, overlapping windows with multiple transformations were used to reduce border effects and checkerboard artifacts. 

\subsection{Results and discussion}

The trained FCN was evaluated on a test set consisting of a single SAR scene containing $99$ avalanches, and its performance was benchmarked against the state-of-the-art algorithm at the time, based on traditional signal processing techniques. The FCN achieved an F1 score of $66.6\%$, substantially outperforming the baseline algorithm, which obtained an F1 score of $38.1\%$. Additionally, the intersection over union score for the FCN was $54.3\%$, compared to $33.1\%$ for the baseline. The FCN correctly detected $72$ avalanches while missing $17$. The missed avalanches were primarily small in size, which are typically more challenging to detect. Furthermore, the FCN produced $32$ false positives, some of which were actual avalanches that had been overlooked during manual annotation by experts.

\begin{figure}
    \centering
    \includegraphics[width=\linewidth]{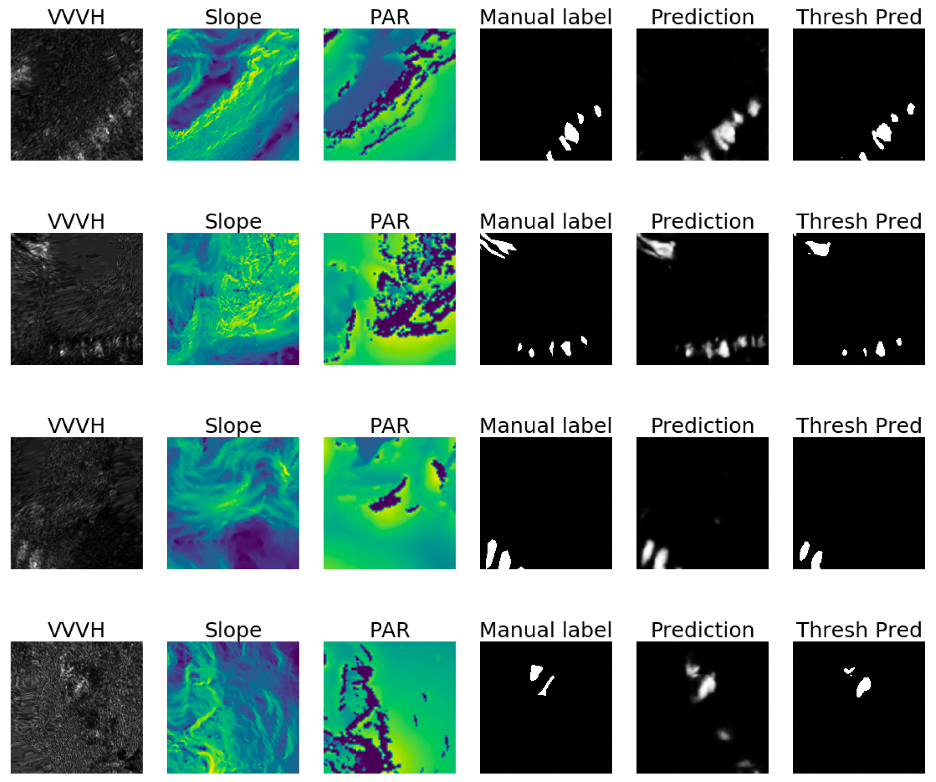}
    \caption{Examples of prediction on individual patches of the validation set. Source: \cite{bianchi2020snow}.}
    \label{fig:fig8}
\end{figure}

Figure~\ref{fig:fig8} presents examples of predictions made by the FCN on patches from the validation set. The figure shows the input SAR channels (VVVH), slope feature, and PAR feature fed into the network, along with the ground truth labels, the raw FCN output, and the thresholded binary output. The predictions demonstrate that the FCN effectively identifies avalanche regions with high accuracy while maintaining sharp boundaries around the detected areas, even for challenging cases with overlapping or clustered avalanches.

\begin{figure}
    \centering
    \includegraphics[width=\linewidth]{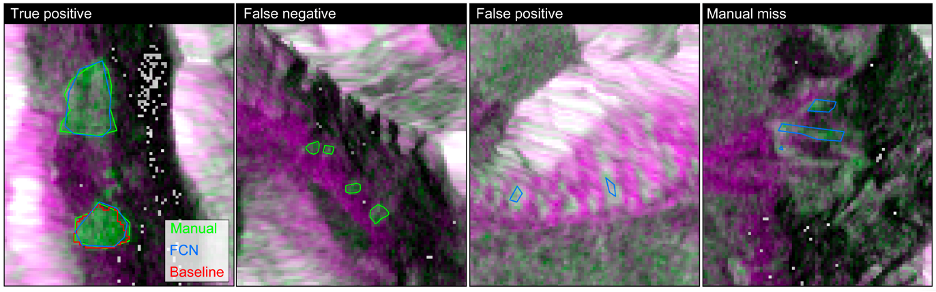}
    \caption{Comparison between manual labeling and FCN output overlain onto a RGB change detection image. Source: \cite{bianchi2020snow}.}
    \label{fig:fig9}
\end{figure}

Figure~\ref{fig:fig9} provides a visual comparison between the manual annotations and the output of the FCN overlaid onto a change detection image. From left to right, the figure highlights: 
(i) areas where the FCN detections match the manual annotations, 
(ii) avalanches missed by the FCN, 
(iii) false detections produced by the FCN, and 
(iv) avalanches correctly detected by the FCN but not marked in the manual annotation. This visual analysis underscores the FCN's potential to detect avalanches that were missed during manual annotation, suggesting that the model can capture subtle patterns in the SAR data that may be overlooked by human experts.

An ablation study was conducted to assess the impact of different input features on the segmentation performance. The study revealed that the inclusion of both VV and VH polarization differences, as well as the derived topographical features (slope and PAR), significantly contributed to the model's accuracy. Moreover, the attention mechanism based on the PAR feature provided further improvements by helping the model focus on the most relevant areas.

In summary, the results showed that the proposed deep learning approach offers a substantial advancement in snow avalanche detection from SAR images, reducing false positives and capturing avalanches that are smaller and more challenging to detect. The integration of topographical features and an attention mechanism into the FCN framework enhanced its detection performance, highlighting the potential for further improvements in operational avalanche monitoring systems.

\section{Improvements and large-scale application}

Recently, we extended the work presented in \cite{bianchi2020snow} by expanding the original dataset and training a wider range of state-of-the-art segmentation architectures and backbones. 
Furthermore, we used one of the newly trained models to generate a large-scale segmentation of snow avalanches across the entirety of Norway, covering the winter seasons from 2016 to 2021. The details of this work extension are discussed in the following.

\subsection{Dataset extension}

A high-quality, comprehensive dataset is crucial to train effective deep learning models.
Therefore, to enhance the segmentation model's performance, we significantly expanded the original dataset. 
This extension allows for training models with higher capacity and improves the performance of previous architectures. 

The updated dataset consists of $4,507$ annotated SAR images, each covering approximately $3.6 \times 3.6 km$, collected over the winter seasons from 2016 to 2020, when both Sentinel-1 A and B satellites were operational. 
The input data includes co-polarized backscatter images (VV and VH channels) from Sentinel-1, captured at two consecutive times from the same orbit, and detailed topographical information, such as slope angles derived from a high-resolution DEM. 
This expanded dataset provides a broader and more diverse set of avalanches across different regions, encompassing a wider range of snow conditions, terrain types, and temporal variations, which can enhance the model's ability to generalize across different environments.

\subsection{Evaluation of State-of-the-art segmentation models}

In recent years, there have been significant advances in deep learning for computer vision, particularly with the development of novel architectures such as Vision Transformers (ViT)~\cite{dosovitskiy2021an}. These models have shown considerable promise in many tasks across various domains. Given this progress, we aimed to evaluate these new approaches in the context of avalanche segmentation, alongside more established convolutional models.

The newly extended dataset was used to train a wide range of state-of-the-art segmentation models in addition to the FCN presented in \cite{bianchi2020snow}. The goal was twofold: first, to identify if there was a more efficient and lightweight segmentation model that could achieve similar or superior performance. Second, to explore the potential of new segmentation models based on ViT, which are more data-hungry.

We tested several well-known architectures, including the U-Net~\cite{ronneberger2015u}, Feature Pyramid Network (FPN)~\cite{lin2017feature}, Pyramid Scene Parsing Network~\cite{zhao2017pyramid}, and DeepLabV3+~\cite{chen2018encoder}. We equipped these architectures with various backbones, including ResNet~\cite{he2016deep}, ResNeSt~\cite{zhang2022resnest}, Res2Ne(X)t~\cite{gao2019res2net}, Xception~\cite{chollet2017xception}, and DenseNet~\cite{huang2017densely}. 
The backbones were initialized with pre-trained weights from ImageNet to leverage transfer learning. The models were implemented with Segmentation Models PyTorch, a framework for testing and benchmarking different architectures.
In addition, we tested a modern U-Net variant adapted from those used in diffusion models~\cite{song2020score}. For Transformer-based architectures, we experimented with UnetFormer~\cite{wang2022unetformer}, SegFormer~\cite{xie2021segformer}, and Dense Prediction Transformer (DPT)~\cite{ranftl2021vision}. 
Finally, we tested the DPT architecture, with various backbones, including Swin Transformer~\cite{liu2021swin}, ConvNeXt~\cite{liu2022convnet}, and ConvNeXtV2~\cite{woo2023convnext}.

Perhaps surprisingly, the best-performing architecture in terms of detection performance and computational complexity was the FPN with an Xception backbone. This result aligns with recent findings by \citep{goldblum2024battle}, which suggest that, despite the recent popularity of ViT, traditional CNNs continue to achieve state-of-the-art performance in many computer vision tasks, including image segmentation.

\subsection{Large-scale detection of avalanches in Norway}

The best-performing model, an FPN with Xception backbone, was used to detect avalanche debris across the whole of Norway from Sentinel-1 SAR data from 2016 to 2020. The input data had a spatial resolution of 10 meters and a 6-day revisit time. To refine the results, filtering criteria were used based on segment area, elevation, and overlap with specific land use types and runout zones.

\begin{figure} 
\centering 
\includegraphics[width=\linewidth]{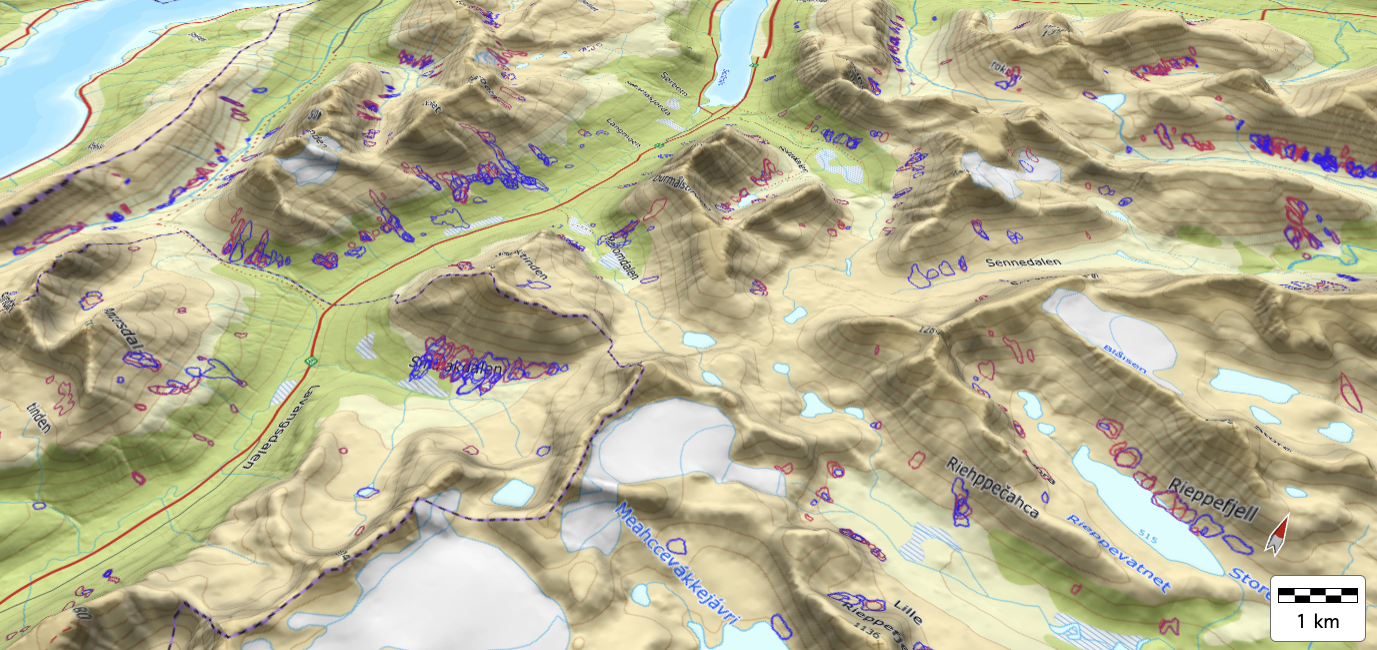} 
\caption{Detected avalanche debris in Troms County, Norway, during a three-month period in 2017. Each polygon represents a detected avalanche, outlining individual debris segments. The visualization is generated using the NLIVE software.} 
\label{fig10} 
\end{figure}

In total, 417,230 avalanche debris segments were detected across Norway. Figure~\ref{fig10} shows the avalanche detections for a three-month period in Troms County. We can see the level of detail of the data, which enables precise analysis of proximity to buildings and roads for example.

\begin{figure} 
\centering 
\includegraphics[width=\linewidth]{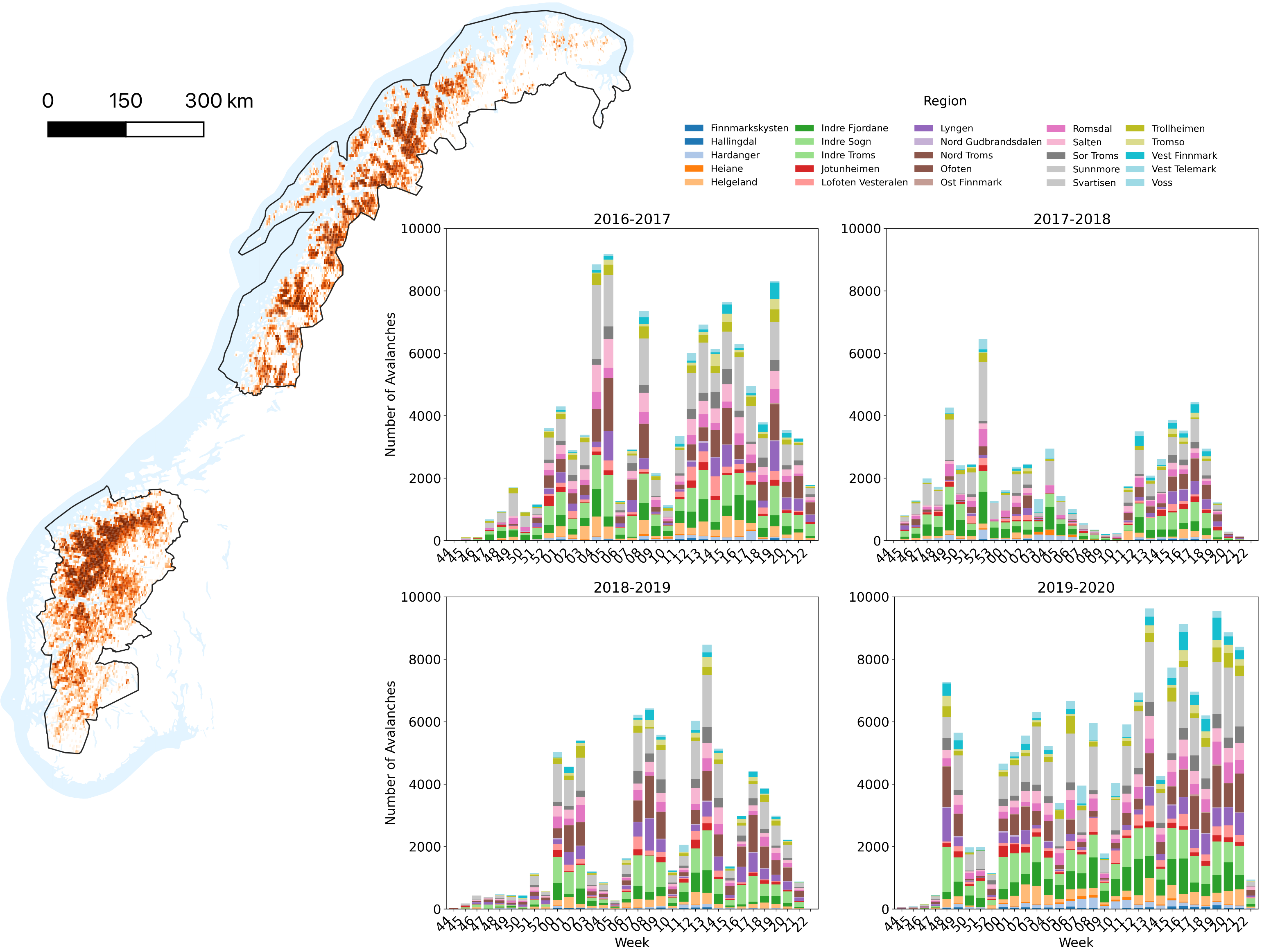} 
\caption{Spatio-temporal distribution of avalanche detections across Norway for the winter seasons 2016-2020. The heat map indicates areas with higher avalanche activity over time. The bar plots show the avalanche count per week, for the four seasons, split into avalanche forecasting regions by their color.} 
\label{fig11} 
\end{figure}

Figure~\ref{fig11} shows the spatial and temporal distribution of the detected avalanches. Such large-scale detection provides insights into avalanche patterns and can be used to improve risk mapping and forecasting efforts. The automated detections also serve as a valuable dataset for training additional models that could compute more accurate risk maps and forecasting.

\section{Conclusion and future work}

Snow avalanches pose significant risks in mountainous regions, and effective monitoring is crucial for public safety and disaster management.
This chapter has provided an overview of the advancements in using deep learning for the detection and segmentation of snow avalanches from SAR data.  Traditional methods for detecting avalanches, such as manual observations and threshold-based signal processing techniques, have proven to be limited in their ability to handle the variability and complexity of SAR data.
The application of deep learning has opened new possibilities for improving avalanche monitoring. The case study by \cite{bianchi2020snow} demonstrated the potential of FCNs for pixel-wise segmentation of avalanches in SAR images, significantly enhancing the detection performance over traditional methods. 

Furthermore, by extending the original dataset and evaluating a wide range of state-of-the-art segmentation models, including ViT, we identified that an FPN with an Xception backbone provides a good tradeoff between performance and computational complexity. 
Our extended dataset, which includes over 4,500 manually annotated SAR images and nearly half a million automated avalanche detections across Norway from 2016 to 2020, serves as a valuable resource for training and testing advanced deep-learning models. 

We plan to leverage this extensive dataset, alongside historical meteorological data and weather forecasts, to predict avalanche activities using spatio-temporal deep learning models. This ongoing effort aims to integrate avalanche detection data with meteorological conditions to enhance forecasting accuracy over large regions where traditional in-situ measurements are sparse or unavailable. Our ultimate goal is to improve real-time avalanche forecasting and provide better tools for risk management in avalanche-prone areas.

\bibliography{Bibliography.bib}

\begin{thebibliography}{41}
\providecommand{\natexlab}[1]{#1}
\providecommand{\url}[1]{\texttt{#1}}
\expandafter\ifx\csname urlstyle\endcsname\relax
  \providecommand{\doi}[1]{doi: #1}\else
  \providecommand{\doi}{doi: \begingroup \urlstyle{rm}\Url}\fi

\bibitem[Alem and Kumar(2020)]{alem2020deep}
A.~Alem and S.~Kumar.
\newblock Deep learning methods for land cover and land use classification in remote sensing: A review.
\newblock In \emph{2020 8th International Conference on Reliability, Infocom Technologies and Optimization (Trends and Future Directions)(ICRITO)}, pages 903--908. IEEE, 2020.

\bibitem[Bianchi et~al.(2020{\natexlab{a}})Bianchi, Espeseth, and Borch]{bianchi2020large}
F.~M. Bianchi, M.~M. Espeseth, and N.~Borch.
\newblock Large-scale detection and categorization of oil spills from sar images with deep learning.
\newblock \emph{Remote Sensing}, 12\penalty0 (14):\penalty0 2260, 2020{\natexlab{a}}.

\bibitem[Bianchi et~al.(2020{\natexlab{b}})Bianchi, Grahn, Eckerstorfer, Malnes, and Vickers]{bianchi2020snow}
F.~M. Bianchi, J.~Grahn, M.~Eckerstorfer, E.~Malnes, and H.~Vickers.
\newblock Snow avalanche segmentation in sar images with fully convolutional neural networks.
\newblock \emph{IEEE Journal of Selected Topics in Applied Earth Observations and Remote Sensing}, 14:\penalty0 75--82, 2020{\natexlab{b}}.

\bibitem[Chen et~al.(2018)Chen, Zhu, Papandreou, Schroff, and Adam]{chen2018encoder}
L.-C. Chen, Y.~Zhu, G.~Papandreou, F.~Schroff, and H.~Adam.
\newblock Encoder-decoder with atrous separable convolution for semantic image segmentation.
\newblock In \emph{Proceedings of the European conference on computer vision (ECCV)}, pages 801--818, 2018.

\bibitem[Chollet(2017)]{chollet2017xception}
F.~Chollet.
\newblock Xception: Deep learning with depthwise separable convolutions.
\newblock In \emph{Proceedings of the IEEE conference on computer vision and pattern recognition}, pages 1251--1258, 2017.

\bibitem[Dosovitskiy et~al.(2021)Dosovitskiy, Beyer, Kolesnikov, Weissenborn, Zhai, Unterthiner, Dehghani, Minderer, Heigold, Gelly, Uszkoreit, and Houlsby]{dosovitskiy2021an}
A.~Dosovitskiy, L.~Beyer, A.~Kolesnikov, D.~Weissenborn, X.~Zhai, T.~Unterthiner, M.~Dehghani, M.~Minderer, G.~Heigold, S.~Gelly, J.~Uszkoreit, and N.~Houlsby.
\newblock An image is worth 16x16 words: Transformers for image recognition at scale.
\newblock In \emph{International Conference on Learning Representations}, 2021.

\bibitem[Eckerstorfer and Malnes(2015)]{eckerstorfer2015manual}
M.~Eckerstorfer and E.~Malnes.
\newblock Manual detection of snow avalanche debris using high-resolution radarsat-2 sar images.
\newblock \emph{Cold Regions Science and Technology}, 120:\penalty0 205--218, 2015.

\bibitem[Eckerstorfer et~al.(2016)Eckerstorfer, B{\"u}hler, Frauenfelder, and Malnes]{eckerstorfer2016remote}
M.~Eckerstorfer, Y.~B{\"u}hler, R.~Frauenfelder, and E.~Malnes.
\newblock Remote sensing of snow avalanches: Recent advances, potential, and limitations.
\newblock \emph{Cold Regions Science and Technology}, 121:\penalty0 126--140, 2016.

\bibitem[Eckerstorfer et~al.(2022)Eckerstorfer, Oterhals, M{\"u}ller, Malnes, Grahn, Langeland, and Velsand]{eckerstorfer2022performance}
M.~Eckerstorfer, H.~D. Oterhals, K.~M{\"u}ller, E.~Malnes, J.~Grahn, S.~Langeland, and P.~Velsand.
\newblock Performance of manual and automatic detection of dry snow avalanches in sentinel-1 sar images.
\newblock \emph{Cold Regions Science and Technology}, 198:\penalty0 103549, 2022.

\bibitem[Fuchs and Bründl(2005)]{fuchs2005}
S.~Fuchs and M.~Bründl.
\newblock Damage potential and losses resulting from snow avalanches in settlements of the canton of grisons, switzerland.
\newblock \emph{Natural Hazards}, 34:\penalty0 53--69, 01 2005.
\newblock \doi{10.1007/s11069-004-0784-y}.

\bibitem[Fuchs et~al.(2007)Fuchs, Th{\"o}ni, McAlpin, Gruber, and Br{\"u}ndl]{fuchs2007}
S.~Fuchs, M.~Th{\"o}ni, M.~C. McAlpin, U.~Gruber, and M.~Br{\"u}ndl.
\newblock Avalanche hazard mitigation strategies assessed by cost effectiveness analyses and cost benefit analyses---evidence from davos, switzerland.
\newblock \emph{Natural Hazards}, 41\penalty0 (1):\penalty0 113--129, 2007.
\newblock \doi{10.1007/s11069-006-9031-z}.
\newblock URL \url{https://doi.org/10.1007/s11069-006-9031-z}.

\bibitem[Gao et~al.(2019)Gao, Cheng, Zhao, Zhang, Yang, and Torr]{gao2019res2net}
S.-H. Gao, M.-M. Cheng, K.~Zhao, X.-Y. Zhang, M.-H. Yang, and P.~Torr.
\newblock Res2net: A new multi-scale backbone architecture.
\newblock \emph{IEEE transactions on pattern analysis and machine intelligence}, 43\penalty0 (2):\penalty0 652--662, 2019.

\bibitem[Gazzea et~al.(2021)Gazzea, Pacevicius, Dammann, Sapronova, Lunde, and Arghandeh]{gazzea2021automated}
M.~Gazzea, M.~Pacevicius, D.~O. Dammann, A.~Sapronova, T.~M. Lunde, and R.~Arghandeh.
\newblock Automated power lines vegetation monitoring using high-resolution satellite imagery.
\newblock \emph{IEEE Transactions on Power Delivery}, 37\penalty0 (1):\penalty0 308--316, 2021.

\bibitem[Geng et~al.(2020)Geng, Jiang, and Deng]{geng2020multi}
J.~Geng, W.~Jiang, and X.~Deng.
\newblock Multi-scale deep feature learning network with bilateral filtering for sar image classification.
\newblock \emph{ISPRS Journal of Photogrammetry and Remote Sensing}, 167:\penalty0 201--213, 2020.

\bibitem[Goldblum et~al.(2024)Goldblum, Souri, Ni, Shu, Prabhu, Somepalli, Chattopadhyay, Ibrahim, Bardes, Hoffman, et~al.]{goldblum2024battle}
M.~Goldblum, H.~Souri, R.~Ni, M.~Shu, V.~Prabhu, G.~Somepalli, P.~Chattopadhyay, M.~Ibrahim, A.~Bardes, J.~Hoffman, et~al.
\newblock Battle of the backbones: A large-scale comparison of pretrained models across computer vision tasks.
\newblock \emph{Advances in Neural Information Processing Systems}, 36, 2024.

\bibitem[Hamar et~al.(2016)Hamar, Salberg, and Ardelean]{hamar2016automatic}
J.~B. Hamar, A.-B. Salberg, and F.~Ardelean.
\newblock Automatic detection and mapping of avalanches in sar images.
\newblock In \emph{2016 IEEE International Geoscience and Remote Sensing Symposium (IGARSS)}, pages 689--692. IEEE, 2016.

\bibitem[He et~al.(2016)He, Zhang, Ren, and Sun]{he2016deep}
K.~He, X.~Zhang, S.~Ren, and J.~Sun.
\newblock Deep residual learning for image recognition.
\newblock In \emph{Proceedings of the IEEE conference on computer vision and pattern recognition}, pages 770--778, 2016.

\bibitem[Huang et~al.(2017)Huang, Liu, Van Der~Maaten, and Weinberger]{huang2017densely}
G.~Huang, Z.~Liu, L.~Van Der~Maaten, and K.~Q. Weinberger.
\newblock Densely connected convolutional networks.
\newblock In \emph{Proceedings of the IEEE conference on computer vision and pattern recognition}, pages 4700--4708, 2017.

\bibitem[Jenkins et~al.(2000)Jenkins, Ranney, Walsh, Miyares, Hewitt, and Collins]{jenkins2000evaluating}
T.~F. Jenkins, T.~A. Ranney, M.~E. Walsh, P.~H. Miyares, A.~D. Hewitt, and N.~H. Collins.
\newblock Evaluating the use of snow-covered ranges to estimate the explosives residues that result from detonation of army munitions.
\newblock 2000.

\bibitem[Kapper et~al.(2023)Kapper, Goelles, Muckenhuber, Tr{\"u}gler, Abermann, Schlager, Gaisberger, Eckerstorfer, Grahn, Malnes, et~al.]{kapper2023automated}
K.~L. Kapper, T.~Goelles, S.~Muckenhuber, A.~Tr{\"u}gler, J.~Abermann, B.~Schlager, C.~Gaisberger, M.~Eckerstorfer, J.~Grahn, E.~Malnes, et~al.
\newblock Automated snow avalanche monitoring for austria: State of the art and roadmap for future work.
\newblock \emph{Frontiers in Remote Sensing}, 4:\penalty0 1156519, 2023.

\bibitem[Kummervold et~al.(2018)Kummervold, Malnes, Eckerstorfer, Arntzen, and Bianchi]{kummervold2018avalanche}
P.~E. Kummervold, E.~Malnes, M.~Eckerstorfer, I.~M. Arntzen, and F.~Bianchi.
\newblock Avalanche detection in sentinel-1 radar images using convolutional neural networks.
\newblock In \emph{Proc. Int. Snow Sci. Workshop}, pages 377--381, 2018.

\bibitem[Lin et~al.(2017)Lin, Doll{\'a}r, Girshick, He, Hariharan, and Belongie]{lin2017feature}
T.-Y. Lin, P.~Doll{\'a}r, R.~Girshick, K.~He, B.~Hariharan, and S.~Belongie.
\newblock Feature pyramid networks for object detection.
\newblock In \emph{Proceedings of the IEEE conference on computer vision and pattern recognition}, pages 2117--2125, 2017.

\bibitem[Linardos et~al.(2022)Linardos, Drakaki, Tzionas, and Karnavas]{linardos2022machine}
V.~Linardos, M.~Drakaki, P.~Tzionas, and Y.~L. Karnavas.
\newblock Machine learning in disaster management: recent developments in methods and applications.
\newblock \emph{Machine Learning and Knowledge Extraction}, 4\penalty0 (2), 2022.

\bibitem[Liu et~al.(2021)Liu, Lin, Cao, Hu, Wei, Zhang, Lin, and Guo]{liu2021swin}
Z.~Liu, Y.~Lin, Y.~Cao, H.~Hu, Y.~Wei, Z.~Zhang, S.~Lin, and B.~Guo.
\newblock Swin transformer: Hierarchical vision transformer using shifted windows.
\newblock In \emph{Proceedings of the IEEE/CVF international conference on computer vision}, pages 10012--10022, 2021.

\bibitem[Liu et~al.(2022)Liu, Mao, Wu, Feichtenhofer, Darrell, and Xie]{liu2022convnet}
Z.~Liu, H.~Mao, C.-Y. Wu, C.~Feichtenhofer, T.~Darrell, and S.~Xie.
\newblock A convnet for the 2020s.
\newblock In \emph{Proceedings of the IEEE/CVF conference on computer vision and pattern recognition}, pages 11976--11986, 2022.

\bibitem[Liu et~al.(2024)Liu, Zhu, Pang, Fu, Zhu, and Liu]{liu2024ava}
Z.~Liu, X.~Zhu, L.~Pang, X.~Fu, H.~Zhu, and X.~Liu.
\newblock Ava-yolo: Image-based multiscale feature fusion enhanced perception model for snow avalanche detection.
\newblock \emph{Measurement Science and Technology}, 2024.

\bibitem[Long et~al.(2015)Long, Shelhamer, and Darrell]{long2015fully}
J.~Long, E.~Shelhamer, and T.~Darrell.
\newblock Fully convolutional networks for semantic segmentation.
\newblock In \emph{Proceedings of the IEEE conference on computer vision and pattern recognition}, pages 3431--3440, 2015.

\bibitem[Luppino et~al.(2021)Luppino, Kampffmeyer, Bianchi, Moser, Serpico, Jenssen, and Anfinsen]{luppino2021deep}
L.~T. Luppino, M.~Kampffmeyer, F.~M. Bianchi, G.~Moser, S.~B. Serpico, R.~Jenssen, and S.~N. Anfinsen.
\newblock Deep image translation with an affinity-based change prior for unsupervised multimodal change detection.
\newblock \emph{IEEE Transactions on Geoscience and Remote Sensing}, 60:\penalty0 1--22, 2021.

\bibitem[Malnes et~al.(2015)Malnes, Eckerstorfer, and Vickers]{malnes2015first}
E.~Malnes, M.~Eckerstorfer, and H.~Vickers.
\newblock First sentinel-1 detections of avalanche debris.
\newblock \emph{The Cryosphere Discussions}, 9\penalty0 (2):\penalty0 1943--1963, 2015.

\bibitem[Ranftl et~al.(2021)Ranftl, Bochkovskiy, and Koltun]{ranftl2021vision}
R.~Ranftl, A.~Bochkovskiy, and V.~Koltun.
\newblock Vision transformers for dense prediction.
\newblock In \emph{Proceedings of the IEEE/CVF international conference on computer vision}, pages 12179--12188, 2021.

\bibitem[Ronneberger et~al.(2015)Ronneberger, Fischer, and Brox]{ronneberger2015u}
O.~Ronneberger, P.~Fischer, and T.~Brox.
\newblock U-net: Convolutional networks for biomedical image segmentation.
\newblock In \emph{Medical image computing and computer-assisted intervention--MICCAI 2015: 18th international conference, Munich, Germany, October 5-9, 2015, proceedings, part III 18}, pages 234--241. Springer, 2015.

\bibitem[Schweizer et~al.(2021)Schweizer, Bartelt, and van Herwijnen]{schweizer2021snow}
J.~Schweizer, P.~Bartelt, and A.~van Herwijnen.
\newblock Snow avalanches.
\newblock In \emph{Snow and ice-related hazards, risks, and disasters}, pages 377--416. Elsevier, 2021.

\bibitem[Song et~al.(2020)Song, Sohl-Dickstein, Kingma, Kumar, Ermon, and Poole]{song2020score}
Y.~Song, J.~Sohl-Dickstein, D.~P. Kingma, A.~Kumar, S.~Ermon, and B.~Poole.
\newblock Score-based generative modeling through stochastic differential equations.
\newblock \emph{arXiv preprint arXiv:2011.13456}, 2020.

\bibitem[Tong et~al.(2024)Tong, Yuan, Zhang, Wang, and Li]{tong2024real}
H.~Tong, J.~Yuan, J.~Zhang, H.~Wang, and T.~Li.
\newblock Real-time wildfire monitoring using low-altitude remote sensing imagery.
\newblock \emph{Remote Sensing}, 16\penalty0 (15):\penalty0 2827, 2024.

\bibitem[Vickers et~al.(2016)Vickers, Eckerstorfer, Malnes, Larsen, and Hindberg]{vickers2016method}
H.~Vickers, M.~Eckerstorfer, E.~Malnes, Y.~Larsen, and H.~Hindberg.
\newblock A method for automated snow avalanche debris detection through use of synthetic aperture radar (sar) imaging.
\newblock \emph{Earth and Space Science}, 3\penalty0 (11):\penalty0 446--462, 2016.

\bibitem[Wang et~al.(2022)Wang, Li, Zhang, Fang, Duan, Meng, and Atkinson]{wang2022unetformer}
L.~Wang, R.~Li, C.~Zhang, S.~Fang, C.~Duan, X.~Meng, and P.~M. Atkinson.
\newblock Unetformer: A unet-like transformer for efficient semantic segmentation of remote sensing urban scene imagery.
\newblock \emph{ISPRS Journal of Photogrammetry and Remote Sensing}, 190:\penalty0 196--214, 2022.

\bibitem[Woo et~al.(2023)Woo, Debnath, Hu, Chen, Liu, Kweon, and Xie]{woo2023convnext}
S.~Woo, S.~Debnath, R.~Hu, X.~Chen, Z.~Liu, I.~S. Kweon, and S.~Xie.
\newblock Convnext v2: Co-designing and scaling convnets with masked autoencoders.
\newblock In \emph{Proceedings of the IEEE/CVF Conference on Computer Vision and Pattern Recognition}, pages 16133--16142, 2023.

\bibitem[Wu et~al.(2023)Wu, Zhang, Xiong, Zhang, Tang, Li, An, and Li]{wu2023near}
X.~Wu, Z.~Zhang, S.~Xiong, W.~Zhang, J.~Tang, Z.~Li, B.~An, and R.~Li.
\newblock A near-real-time flood detection method based on deep learning and sar images.
\newblock \emph{Remote Sensing}, 15\penalty0 (8):\penalty0 2046, 2023.

\bibitem[Xie et~al.(2021)Xie, Wang, Yu, Anandkumar, Alvarez, and Luo]{xie2021segformer}
E.~Xie, W.~Wang, Z.~Yu, A.~Anandkumar, J.~M. Alvarez, and P.~Luo.
\newblock Segformer: Simple and efficient design for semantic segmentation with transformers.
\newblock \emph{Advances in neural information processing systems}, 34:\penalty0 12077--12090, 2021.

\bibitem[Zhang et~al.(2022)Zhang, Wu, Zhang, Zhu, Lin, Zhang, Sun, He, Mueller, Manmatha, et~al.]{zhang2022resnest}
H.~Zhang, C.~Wu, Z.~Zhang, Y.~Zhu, H.~Lin, Z.~Zhang, Y.~Sun, T.~He, J.~Mueller, R.~Manmatha, et~al.
\newblock Resnest: Split-attention networks.
\newblock In \emph{Proceedings of the IEEE/CVF conference on computer vision and pattern recognition}, pages 2736--2746, 2022.

\bibitem[Zhao et~al.(2017)Zhao, Shi, Qi, Wang, and Jia]{zhao2017pyramid}
H.~Zhao, J.~Shi, X.~Qi, X.~Wang, and J.~Jia.
\newblock Pyramid scene parsing network.
\newblock In \emph{Proceedings of the IEEE conference on computer vision and pattern recognition}, pages 2881--2890, 2017.

\end{thebibliography}

\end{document}